\documentclass[10pt]{bytedance_seed}
\usepackage{amsmath,amssymb,array,tabularx,tikz}
\usetikzlibrary{arrows.meta,positioning}
\title{GUI-WEdit and GUI-SkillEvo:\\Gradient-Free Adaptation of Small GUI Agents}

\author[1]{Zhengyong Wan}
\author[1]{Kai Fu}
\author[1]{Yuzhen Tong}
\author[2]{Yingtian Chen}
\author[2]{Wei Dai}
\addtolist[2]{Sihan Yang}{\authorlist}{\authorformat}{\\}
\author[2]{Zuzheng Kuang}
\author[2]{Haixia Bi}
\author[3]{Jian Sun}
\affiliation[1]{Ninetech Information Technology (Shenzhen) Co., Ltd.}
\affiliation[2]{School of Information and Communications Engineering, Xi'an Jiaotong University, Xi'an 710049, China}
\affiliation[3]{School of Mathematics and Statistics, Xi'an Jiaotong University}

\affiliation{\href{mailto:haixia.bi@xjtu.edu.cn}{\texttt{haixia.bi@xjtu.edu.cn}}}
\hypersetup{pdfauthor={Haixia Bi, Yingtian Chen, Sihan Yang, Zuzheng Kuang, Yuzhen Tong, Kai Fu, Zhengyong Wan}}
\contribution{Technical report --- November 2026}
\abstract{Small graphical user interface (GUI) agents can execute desktop workflows, but repeated local errors and fragile procedural decisions limit their reliability. We present GUI-WEdit-4B and GUI-SkillEvo-2B, two gradient-free adaptations of GUI-Owl-1.5 backbones. For the 4B model, we use trajectory feedback to construct guarded, closed-form updates to a late multilayer perceptron (MLP), retaining successful states as protection constraints. The final checkpoint changes 23,739,095 parameter values, or 0.535\% of the 4,437,815,808-parameter backbone, within one weight tensor. For the 2B model, an assistant iteratively proposes and refines procedural skills while the model weights remain fixed. On the 361-task OSWorld subset excluding Google Drive tasks, the frozen systems obtain final aggregate scores of 48.02\% and 45.47\%, respectively.}
\date{November 27, 2026}
\begin{document}
\maketitle

\section{Introduction}
OSWorld evaluates agents through executable desktop tasks and task-specific evaluators, requiring a model to translate visual observations into actions across applications~\citep{osworld}. A small model may locate the correct interface element yet fail by selecting an inappropriate mode, repeating a completed operation, or terminating at the wrong time. These failures motivate adaptation at two different points: the policy weights that produce actions and the procedural context supplied to the policy.

We investigate these routes using GUI-Owl-1.5-4B-Instruct and GUI-Owl-1.5-2B-Instruct~\citep{guiowl}. We call the resulting systems GUI-WEdit-4B (weight editing) and GUI-SkillEvo-2B (skill evolution). The 4B route edits a small fraction of the existing parameters without backpropagation or an optimizer. The 2B route keeps the backbone fixed and refines skills through execution feedback. Both retain the OSWorld task definitions and scoring criteria. Our contribution is an engineering study of a feedback-driven adaptation workflow, an audited 4B checkpoint with localized weight changes, and a frozen 2B skill harness with documented execution settings.

Weight-space programming motivates using a language-model assistant to inspect failures and implement targeted changes to model parameters~\citep{wsp,robotedit}. Our implementation uses supervised state--action evidence and a constrained closed-form solve. For skills, we adopt the propose--evaluate--retain principle associated with recursive self-improvement~\citep{rsi}. Agent S2 studies a compositional computer-use harness~\citep{agents2,li2026dynamicagentskillslifecycle}.

\paragraph{Related visual modeling.}
In remote sensing, methods incorporate task-specific structure through three-dimensional wavelet textures with Markov random fields or through unsupervised polarimetric image factorization with convolutional networks~\citep{bi2017polsarwavelet,bi2019polsarfactorization}. More recent work uses adaptive graph relations for multilabel scene classification, local enhancement and cross-modal state-space fusion for land-cover classification, and multiscale contrastive pretraining for polarimetric image classification~\citep{bi2024crossattention,chang2026lsfmamba,kuang2025ecpmamba}. In a different domain, Transformers have been applied to the inverse design of terahertz polarization converters~\citep{mao2024terahertz}. These studies adapt representations or architectures to their domains. Here, we instead start with existing GUI policies and adapt their behavior through localized weight edits or procedural guidance, without retraining their visual backbones.

\section{Feedback-Driven Adaptation}
\label{sec:method}
\begin{figure}[t]
\centering
\includegraphics[width=\linewidth]{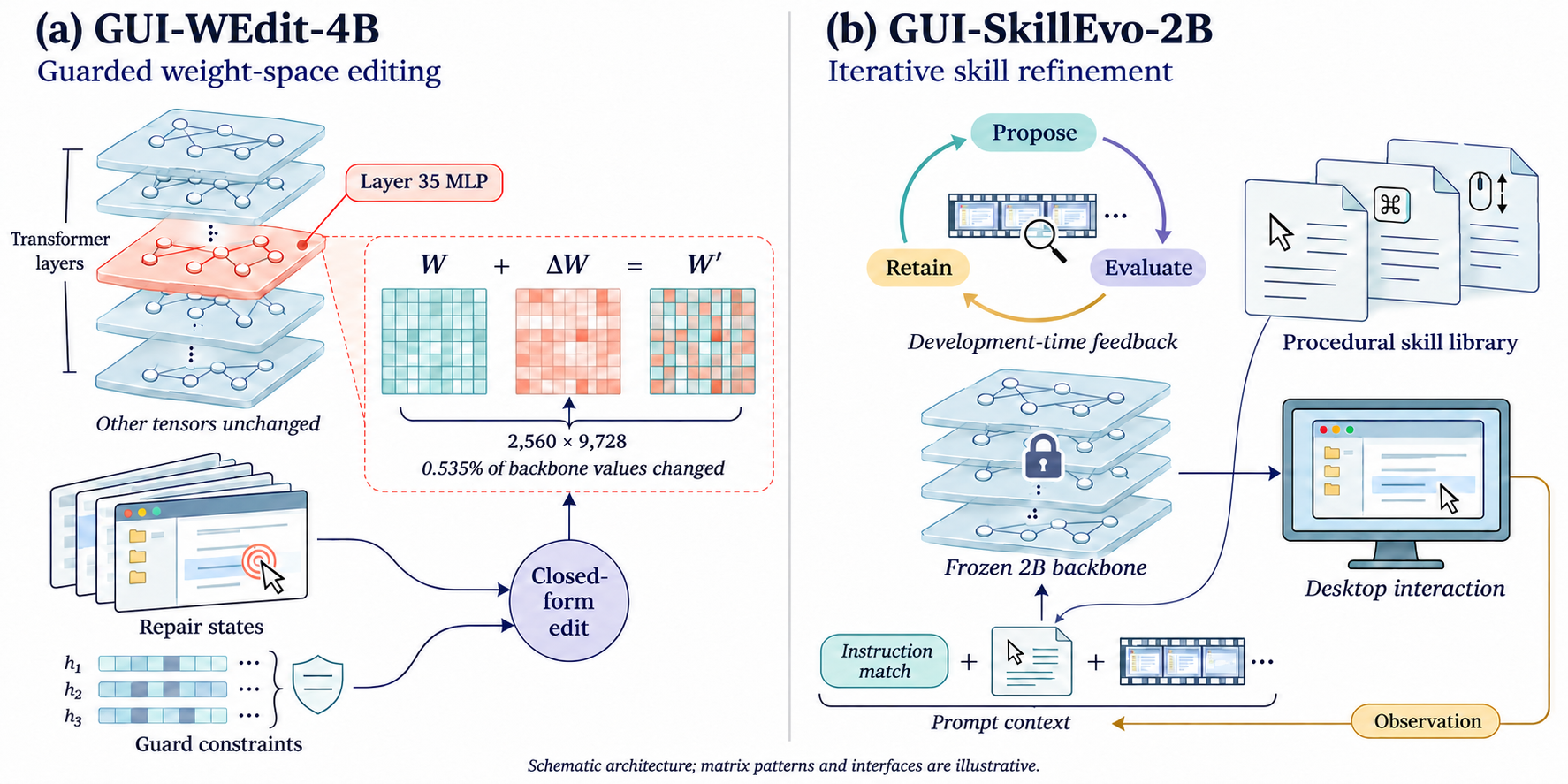}
\caption{Development and execution of the two adaptation routes. Repair and guard evidence inform the 4B closed-form edit; execution feedback refines the 2B skill library. The accepted checkpoint and skill library are fixed for evaluation. Matrix tiles and the GUI thumbnail are schematic.}
\label{fig:workflow}
\end{figure}

Development uses feedback from public OSWorld tasks. We inspect recorded screenshots, action sequences, and evaluator outputs to identify actionable failures. Candidate changes are checked against successful reference states and then evaluated in closed-loop desktop interaction. The research assistant operates offline during development. Figure~\ref{fig:workflow} summarizes the two routes.

\subsection{GUI-WEdit-4B: Guarded Weight-Space Editing}
\paragraph{Edit location.}
The selected matrix is \texttt{model.language\_model.layers.35.mlp.down\_proj.weight}, using zero-based layer numbering. It maps a 9,728-dimensional MLP activation to the 2,560-dimensional residual stream. Let this matrix be $W\in\mathbb{R}^{2560\times9728}$. We save a separate checkpoint with $W'=W+\Delta W$. Vision parameters, token embeddings, attention weights, other MLP layers, and model configuration are unchanged.

\paragraph{Target construction and protection.}
A desired action is chosen from the diagnosed interaction state. With the instruction, screenshots, history, and preceding response tokens fixed, we capture MLP inputs and residual activations using the serving implementation. Desired token readout margins define residual corrections $U$ at target activation columns $K$. Successful reference states form a guard bank. A compact form of the construction objective is
\begin{equation}
\min_{\Delta W}\; \sum_{j\in\mathcal{G}}\frac{\|\Delta W g_j\|_2^2}{\|h_j\|_2^2}
+\lambda\|\Delta W D^{-1}\|_F^2,
\qquad \text{subject to }\Delta W K=U,
\label{eq:edit}
\end{equation}
where $g_j$ is a guard activation, $h_j$ its residual state, and $D$ contains the original column norms. The normalization discourages large relative changes to successful states. Writing the resulting positive-definite geometry as $C$, the equality-constrained solution is
\begin{equation}
\Delta W=U(K^{\mathsf T}C^{-1}K)^{-1}K^{\mathsf T}C^{-1}.
\end{equation}
Hard protection anchors are represented by zero-correction columns in $U$. We bound relative column and matrix perturbations, account for BF16 rounding and the configured repetition penalty in token checks, and reject candidates that violate the declared checks after serialization. These are analytic interventions, with gradients disabled and no optimizer updates. Guard residual perturbation screens candidates before closed-loop evaluation.

\paragraph{Final artifact.}
The final construction stage used one target input and 812 guard inputs, yielding 34,707 protected token states, including already-correct positions from the target response. It applied one corrective target position and one hard zero-change anchor. Earlier accepted edits are retained in the final checkpoint. An elementwise audit against the original backbone found 23,739,095 changed values in this single tensor out of 4,437,815,808 total parameter values (0.5349\%). The other 712 serialized tensors are identical. No skill text is injected in the 4B evaluation.

\subsection{GUI-SkillEvo-2B: Iterative Skill Refinement}
The second route iteratively refines procedural skills while keeping the 2B backbone fixed. A development assistant inspects recorded screenshots and actions to diagnose a failed execution, then drafts a skill containing action guidance and stopping conditions. The skill is evaluated through the normal agent interface. The assistant uses the resulting trajectory and task score to revise the procedure or retain it in the skill library.

The accepted library is fixed during evaluation. The wrapper uses exact instruction matching to select a procedure and append it to the instruction-side context; unmatched instructions retain the original prompt. At each step, the agent observes the current screenshot and generates a computer-use action. This separates offline skill refinement from online action generation, with the model weights remaining bitwise unchanged.

\section{Evaluation Setup}
\label{sec:setup}
\paragraph{Task scope and metric.}
We use the same fixed 361 tasks in ten domains, excluding the eight Google Drive tasks from the 369-task OSWorld suite. This subset is supported by the OSWorld setup guide~\citep{osworldguide}. The task scope follows OSWorld 1.0. We preserve raw finite scores from \texttt{result.txt}, including partial credit. A full-score task has score exactly one; the headline metric is the arithmetic mean of task scores, multiplied by 100.

\paragraph{Execution parameters.}
Table~\ref{tab:settings} records the actual settings of the frozen systems. The history parameter retains up to five image observations under the native agent's message construction. Both systems use screenshot observations and the original GUI-Owl action parser connected to OSWorld's parallel runner. The 4B route changes model weights, and the 2B route adds skill context; neither changes evaluator functions, scoring thresholds, or task objectives.

\begin{table}[!htbp]
\centering\small
\caption{Recorded execution settings for GUI-WEdit-4B and GUI-SkillEvo-2B.}
\label{tab:settings}
\begin{tabularx}{\linewidth}{@{}lXX@{}}
\toprule
Setting & GUI-WEdit-4B & GUI-SkillEvo-2B\\
\midrule
Backbone & GUI-Owl-1.5-4B-Instruct & GUI-Owl-1.5-2B-Instruct\\
History parameter & 5 & 5\\
Runner \texttt{max\_steps} & 50 & 75\\
Agent \texttt{max\_trajectory\_length} & 50 & 50\\
Temperature / top-$p$ / top-$k$ & 0 / 0.9 / 1 & 0 / 0.9 / 1\\
Repetition penalty & 1.05 & 1.05\\
Maximum output tokens & 4,096 & 4,096\\
Screen resolution & $1920\times1080$ & $1920\times1080$\\
Post-action wait & 4 seconds & 4 seconds\\
\texttt{keep\_full\_response} & 0 & 0\\
Environments per runner & 2 & 4\\
Serving GPU / tensor parallelism & A100 / 1 & A100 / 1\\
vLLM / PyTorch & 0.30.0 / 2.13.0 & 0.25.1 / 2.11.0\\
Task skill injection & Off & Enabled\\
\bottomrule
\end{tabularx}
\vspace{1mm}
\begin{minipage}{\linewidth}\footnotesize
The native agent returns \texttt{FAIL} when its response-history count reaches \texttt{max\_trajectory\_length}. Both systems therefore retain the agent's 50-response termination rule.
\end{minipage}
\end{table}

\paragraph{Score aggregation.}
We report the arithmetic mean of the valid score for each of the 361 tasks. Historical backbone test scores are included for reference.

\FloatBarrier
\section{Results}
\begin{table}[!htbp]
\centering\small
\caption{Final results and historical project references on the same 361-task scope.}
\label{tab:main}
\begin{tabular}{@{}lrrrr@{}}
\toprule
System & Historical base (\%) & Final (\%) & Difference (pp) & Full score\\
\midrule
GUI-SkillEvo-2B & 37.68 & 45.47 & +7.79 & 159/361\\
GUI-WEdit-4B & 43.43 & 48.02 & +4.59 & 168/361\\
\bottomrule
\end{tabular}
\end{table}
The frozen 4B checkpoint obtains a final aggregate of 48.02\%, and the frozen 2B system obtains 45.47\% (Table~\ref{tab:main}). The 4B result is 4.59 percentage points above its historical project reference, while the 2B difference is 7.79 points.

Table~\ref{tab:domains} gives all ten domains. The 4B system scores 80.77\% on GIMP, 69.57\% on VS Code, and 56.51\% on Writer. The 2B system scores 93.33\% on Thunderbird, 91.67\% on operating-system tasks, and 69.57\% on VS Code. Multi-application workflows and Calc remain difficult for both systems.

\begin{table}[!htbp]
\centering\small
\caption{Domain-level final mean scores and full-score counts. Overall means weight domains by their task counts.}
\label{tab:domains}
\begin{tabular}{@{}lrrrrr@{}}
\toprule
Domain & Tasks & 2B mean (\%) & 2B full & 4B mean (\%) & 4B full\\
\midrule
Chrome & 46 & 43.39 & 19 & 45.56 & 20\\
GIMP & 26 & 69.23 & 18 & 80.77 & 21\\
LibreOffice Calc & 47 & 25.53 & 12 & 27.66 & 13\\
LibreOffice Impress & 47 & 33.84 & 15 & 53.06 & 24\\
LibreOffice Writer & 23 & 65.21 & 14 & 56.51 & 12\\
Multi-application & 93 & 26.23 & 23 & 30.71 & 27\\
Operating system & 24 & 91.67 & 22 & 66.67 & 16\\
Thunderbird & 15 & 93.33 & 14 & 66.67 & 10\\
VLC & 17 & 40.56 & 6 & 58.21 & 9\\
VS Code & 23 & 69.57 & 16 & 69.57 & 16\\
\midrule
Overall & 361 & 45.47 & 159 & 48.02 & 168\\
\bottomrule
\end{tabular}
\end{table}

\paragraph{Task-level outcomes.}
Against the historical 4B reference, 38 tasks improve, 20 decrease, and 303 retain the same score.

\FloatBarrier
\section{Conclusion}
We document two practical routes to adapting small desktop agents without conventional gradient-based model training. Guarded weight-space editing concentrates the 4B changes in one late MLP tensor, while iterative skill refinement adapts the 2B harness with fixed model weights. The frozen systems reach final 361-task aggregates of 48.02\% and 45.47\%.

\bibliographystyle{unsrtnat}
\bibliography{references}
\end{document}